%% file: main.tex
\documentclass[letterpaper, 10 pt, conference]{ieeeconf}  

\usepackage{times}
\usepackage{epsfig}

\usepackage{blindtext}
\usepackage{adjustbox}
\usepackage{caption}
\usepackage[font=small,labelfont=bf]{caption}
\usepackage{capt-of}
\usepackage{hhline}

\usepackage{soul}
\usepackage[normalem]{ulem}

\usepackage[breaklinks=true,bookmarks=false]{hyperref}
\usepackage{cite}
\usepackage{amsmath,amssymb,amsfonts}
\usepackage{algorithmic}
\usepackage{graphicx}
\usepackage{subcaption}
\usepackage{textcomp}
\usepackage{xcolor}
\usepackage{multirow}
\usepackage{authblk}
\usepackage{tabularx}
\usepackage{booktabs}
\usepackage{tikz}
\usepackage{hyperref}
\usepackage{blindtext}

\usepackage{multirow}

\definecolor{darkblue}{rgb}{0.2,0,0.9}
\definecolor{darkorange}{rgb}{0.8,0.4,0}
\definecolor{darkgreen}{rgb}{0,0.6,0}
\definecolor{darkpurple}{rgb}{0.6,0,0.7}
\definecolor{darkteal}{rgb}{0,0.5,1}
\definecolor{deepred}{RGB}{180,0,0}
\definecolor{deepgreen}{RGB}{0,150,0}

\newcommand{\cmark}{\textbf{\textcolor{deepgreen}{\checkmark}}}
\newcommand{\xmark}{\textbf{\textcolor{deepred}{$\times$}}}

\title{ATARS: An Aerial Traffic Atomic Activity Recognition and 
Temporal Segmentation Dataset}

\author{Zihao Chen\\
National Chengchi University\\
{\tt\small 113761501@nccu.edu.tw}

\and
Hsuanyu Wu\\
National Chengchi University\\
{\tt\small 112753109@nccu.edu.tw}

\and
Chi-Hsi Kung$^{\ast}$\\
Indiana University Bloomington\\
{\tt\small kung@iu.edu}

\and
Yi-Ting Chen$^{\ast}$\\
National Yang Ming Chiao Tung University\\
{\tt\small ychen@cs.nycu.edu.tw}

\and
Yan-Tsung Peng$^{\ast}$\\
National Chengchi University\\
{\tt\small ytpeng@cs.nccu.edu.tw}

\thanks{$^{\ast}$ Equal Advising.}
}

\begin{document}

\twocolumn[{%
    \renewcommand\twocolumn[1][]{#1}%
    \maketitle
    \begin{center}
        \centering
        \captionsetup{type=figure}
        \includegraphics[width=1.0\textwidth]{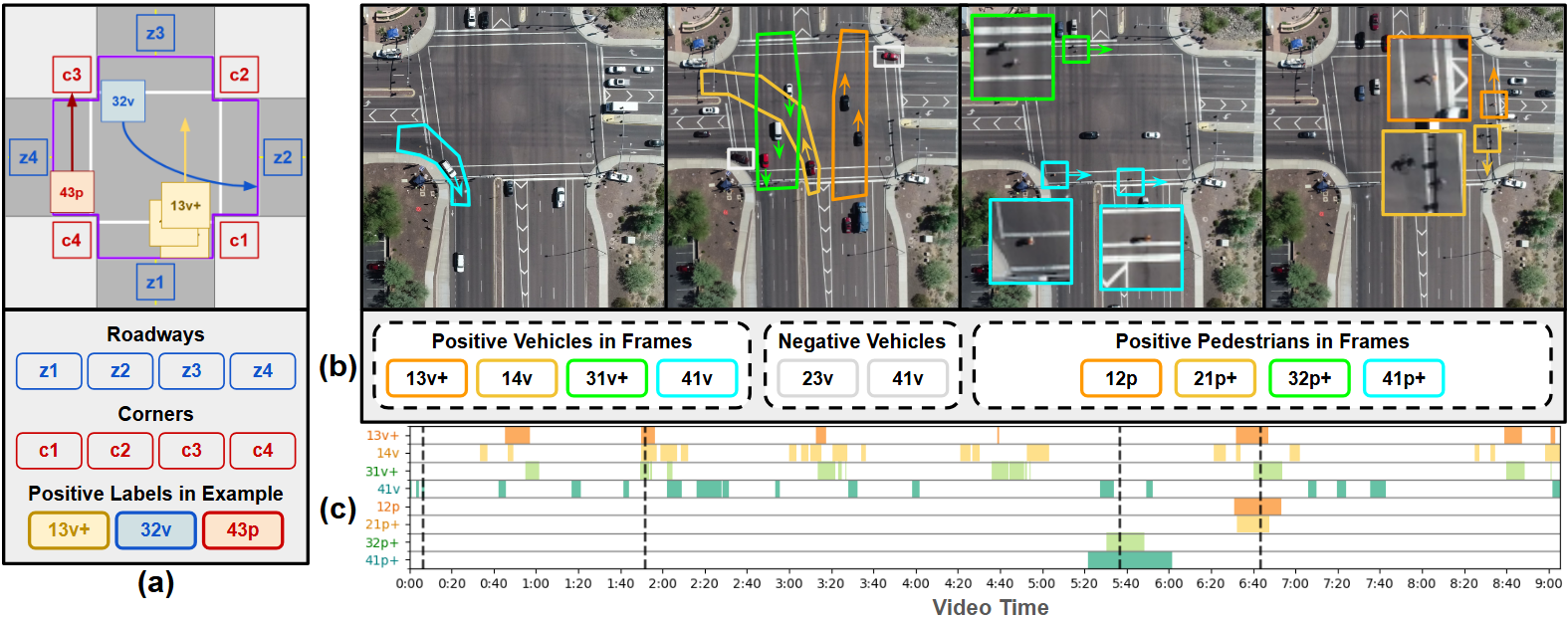} 
        \captionof{figure}{Illustration of the ATARS dataset. (a) Atomic Activity definition showing the names of each roadway (\textbf{z1}, \textbf{z2}, \textbf{z3}, and \textbf{z4}), four corners (\textbf{c1}, \textbf{c2}, \textbf{c3}, and \textbf{c4}), and examples of atomic activity labels. The region boxed in purple is the intersection and all traffic participants moving inside are annotated. Labels containing \textbf{v} and \textbf{p} represent a singular vehicle or pedestrian, respectively. The additional `+' represents multiple vehicles or pedestrians in the intersection. (b) Chronologically ordered frames sampled from a video, highlight atomic activity and the challenge of detecting small objects, such as pedestrians, due to the high-altitude top-down perspective. (c) Visualization of atomic activity labels annotated at frame granularity, with dashed lines marking the location of the presented frames in (b).
            }
    \label{fig:teaser}
    \end{center}
}]

\input{chapters/1_abstract}

\input{chapters/2_intro}

\input{chapters/3_related_work}

\input{chapters/4_methodology}

\input{chapters/5_experiments}

\input{chapters/6_discussion}

\input{chapters/7_conclusion}
\bibliographystyle{IEEEtran}
\bibliography{egbib}

\end{document}

%% file: chapters/1_abstract.tex
\begin{abstract}
Traffic Atomic Activity which describes traffic patterns for topological intersection dynamics is a crucial topic for the advancement of intelligent driving systems. However, existing atomic activity datasets are collected from an egocentric view, which cannot support the scenarios where traffic activities in an entire intersection are required.
Moreover, existing datasets only provide video-level atomic activity annotations, which require exhausting efforts to manually trim the videos for recognition and limit their applications to untrimmed videos.
To bridge this gap, we introduce the Aerial Traffic Atomic Activity Recognition and Segmentation (ATARS) dataset, the first aerial dataset designed for multi-label atomic activity analysis. We offer atomic activity labels for each frame, which accurately record the intervals for traffic activities. 
Moreover, we propose a novel task, Multi-label Temporal Atomic Activity Recognition, enabling the study of accurate temporal localization for atomic activity and easing the burden of manual video trimming for recognition.
We conduct extensive experiments to evaluate existing state-of-the-art models on both atomic activity recognition and temporal atomic activity segmentation. The results highlight the unique challenges of our ATARS dataset, such as recognizing extremely small objects' activities. 
We further provide comprehensive discussion analyzing these challenges and offer valuable insights for future direction to improve recognizing atomic activity in aerial view. Our source code and dataset are available at \href{https://github.com/magecliff96/ATARS/}{\textcolor{magenta}{https://github.com/magecliff96/ATARS/}}.

\end{abstract}


%% file: chapters/2_intro.tex
\section{Introduction}
\label{sec:intro}
\begin{table*}[t!]
\centering
\scriptsize
\caption{Comparison of traffic datasets relevant to traffic activity analysis for intersection dynamics, where RTA denotes Road Topology Activity, \textcolor{darkteal}{AA} denotes atomic activity labels, and \textcolor{darkpurple}{Triplet} refers to the actor-action-location triplet labels used in ROAD~\cite{singh2022road}.}
\resizebox{0.85\linewidth}!{
\begin{tabular}
    {@{} c@{\;} | c |  c@{\;} c@{\;} c@{\;} c@{\;} c@{\;} c@{\;} }
    \toprule
    \multirow{2}{*}{Dataset} &  
    \multirow{2}{*}{Perspective} &  
    \multirow{2}{*}{Real Data} &  
    \multicolumn{1}{c}{Video RTA} &  
    \multicolumn{1}{c}{Frame RTA} &  
    \multicolumn{1}{c}{Pedestrian} &  
    \multirow{2}{*}{Trajectory} &  
    \multicolumn{1}{c}{Bounding}  
    \\  
    &  
    &  
    &  
    \multicolumn{1}{c}{Labels} &  
    \multicolumn{1}{c}{Labels} &  
    \multicolumn{1}{c}{Labels} &  
    &  
    \multicolumn{1}{c}{Boxes}
    \\  
    \midrule
    DRAMA~\cite{malla2023drama} & \textcolor{darkblue}{Egocentric} & \cmark & \xmark & \xmark & \cmark & \xmark & \cmark  
    \\  
    HDD~\cite{ramanishka2018toward} & \textcolor{darkblue}{Egocentric} & \cmark & \xmark & \xmark & \xmark & \xmark & \cmark  
    \\  
    ROAD~\cite{singh2022road} & \textcolor{darkblue}{Egocentric} & \cmark & \xmark & \textcolor{darkpurple}{Triplet} & \cmark & \xmark & \cmark  
    \\  
    OATS~\cite{agarwal2023ordered} & \textcolor{darkblue}{Egocentric} & \cmark & \textcolor{darkteal}{AA} & \xmark & \cmark & \xmark & \xmark  
    \\  
    TACO~\cite{kung2024action} & \textcolor{darkblue}{Egocentric} & \xmark & \textcolor{darkteal}{AA} & \xmark & \cmark & \xmark & \xmark  
    \\  
    
    \midrule
    UAVDT~\cite{du2018unmanned} & \textcolor{darkorange}{Drone} & \cmark & \xmark & \xmark & \cmark & \xmark & \cmark  
    \\  
    VisDrone~\cite{zhu2021detection} & \textcolor{darkorange}{Drone} & \cmark & \xmark & \xmark & \cmark & \cmark & \cmark  
    \\  
    
    \midrule
    CitySim~\cite{zheng2024citysim} & \textcolor{darkgreen}{Top-Down} & \cmark & \xmark & \xmark & \xmark & \cmark & \cmark  
    \\  
    INTERACTION~\cite{zhan2019interaction} &  \textcolor{darkgreen}{Top-Down} & \cmark & \xmark & \xmark & \cmark & \cmark & \xmark  
    \\  
    inD~\cite{bock2020ind} & \textcolor{darkgreen}{Top-Down} & \cmark & \xmark & \xmark & \cmark & \cmark & \cmark  
    \\  
    SinD~\cite{xu2022drone} & \textcolor{darkgreen}{Top-Down}  & \cmark & \xmark & \xmark & \cmark & \cmark & \cmark  
    \\  
    CAROM~\cite{lu2023carom} & \textcolor{darkgreen}{Top-Down} & \cmark & \xmark & \xmark & \xmark & \cmark & \cmark  
    \\  
    \midrule
    ATARS (ours) & \textcolor{darkgreen}{Top-down} & \cmark & \textcolor{darkteal}{AA} & \textcolor{darkteal}{AA} & \cmark & \cmark & \cmark  
    \\  
    \bottomrule
\end{tabular}
}
\label{table:dataset}
\vspace{-5mm}

\end{table*}
Understanding traffic scenes that recognize and model the behavior of vehicles and pedestrians in complex and dynamic traffic scenes plays a crucial role in many applications, including scenario retrieval~\cite{lin2014visual,segal2021universal,naphade20237th,taha2020boosting},
, safety-critical scenariogeneration~\cite{cao2022advdo,Rempe_2022_CVPR,Rempe_2023_CVPR,NEURIPS2022_a48ad12d}, and traffic risk assessment~\cite{make_stop,droid,riskbench,pao2024potential}.
Extensive research efforts have been conducted on visual traffic activity understanding by studying the recognition of high-level actions,  such as proceeding forward and turning right, from egocentric videos~\cite{chen2016atomic,ramanishka2018toward,li2019dbus,malla2020titan}.
More recently, the task of multi-label atomic activity (AA) is introduced, exploiting road topology information to enhance the granularity of activity descriptions~\cite{agarwal2023ordered,kung2024action}.
As depicted in Figure~\ref{fig:teaser} (a), the embedded road topology information allows for finer-grained and machine-friendly action descriptions by defining atomic activities based on movement across different road topological regions and agent categories. This can thus offer a more comprehensive analysis for traffic flow dynamics.
However, existing datasets OATS~\cite{agarwal2023ordered} and TACO~\cite{kung2024action} are limited to an ego-centric perspective, relying on dashboard cameras and vehicle-mounted sensors. Their constrained visibility and ego-motion hinder applications in intersection traffic analysis~\cite{hu2024optimizing} and simulation~\cite{dosovitskiy2017carla}, both of which require video footage with a stable camera position from a top-down view that allows focus on the entire intersection. 

To bridge this gap, we present the first atomic activity dataset collected from a top-down view, the Aerial Traffic Atomic Acitivity Recognition and Segmentation (ATARS) dataset.
Captured from a drone perspective, the dataset is designed to study the dynamic traffic activities of entire intersections, encompassing multiple small agents in the scene.
We follow the atomic activity definition in OATS~\cite{agarwal2023ordered} and TACO~\cite{kung2024action} to annotate ATARS, as depicted in Figure~\ref{fig:teaser} (b). 
The top-down view of the atomic activity dataset opens new doors for traffic video understanding and introduces unique challenges, such as associating road topology context with extremely small traffic participants.

Moreover, unlike existing atomic activity (AA) datasets, which are constrained by their video-level label granularity and thus require curated and trimmed videos, we provide frame-level atomic activity labels that describe the temporal duration of actions for untrimmed videos, and we further propose a novel task: Multi-label Temporal Atomic Activity (M-TAA) Segmentation.
Figure~\ref{fig:teaser} (c) illustrates examples of annotations for temporal dense multi-label prediction within an untrimmed parent video. 
This new task not only enables a more detailed analysis of traffic participants' behaviors but also eases the burden of manual short-clip trimming for AA recognition~\cite{agarwal2023ordered,kung2024action}.

To benchmark the effectiveness of state-of-the-art models on the ATARS, we conduct extensive evaluations using established methods in traffic activity recognition~\cite{baradel2018object, wu2019learning, feichtenhofer2019slowfast, feichtenhofer2020x3d, fan2021multiscale, arnab2021vivit, tong2022videomae, kung2024action} and temporal action segmentation~\cite{farha2019ms, li2020ms, yi2021asformer, tan2022pointtad}. Our results highlight the unique challenges of addressing the long-tailed distribution of real-world data and capturing fine-grained action dynamics of small agents in aerial data. We further provide a comprehensive discussion to analyze these challenges and offer insightful future directions.

In summary, our contributions are threefold.
\begin{enumerate}
    \item We propose the first top-down view dataset, ATARS, designed for multi-label atomic activity, opening new opportunities for research on topology-aware atomic activity recognition and multi-label temporal atomic activity segmentation for intersections.
    \item We offer both video and frame-level atomic activity labels, enabling top-down view atomic activity recognition and multi-label temporal activity segmentation.
    \item Our comprehensive benchmark evaluations with state-of-the-art models provide insightful analyses of the effectiveness of current approaches and highlight the unique challenges posed by small traffic participants, temporal localization, and long-tailed distribution.
\end{enumerate}

%% file: chapters/3_related_work.tex
\section{Related Work}
\label{sec:related}

\subsection{Traffic Activity Recognition Datasets}
Research for intelligent driving systems, such as scene analysis~\cite{agarwal2023ordered,du2018unmanned,zheng2024citysim,zhan2019interaction,xu2022drone,lu2023carom} and scenario retrieval~\cite{agarwal2023ordered,malla2023drama,lu2023carom}, heavily relies on traffic activity understanding. Extensive early traffic datasets focus on object-level detection and tracking~\cite{geiger2013vision,yu2020bdd100k,caesar2020nuscenes} and trajectory prediction~\cite{chang2019argoverse,sun2020scalability}. Recently, several datasets have framed traffic activity understanding as action recognition and detection tasks~\cite{chen2016atomic,li2019dbus,ramanishka2018toward,malla2020titan,riskbench}.
For example, 
HDD~\cite{ramanishka2018toward} datasets provide high-level action labels (e.g., left turn) to describe dynamic scenes. However, these labels lack essential topological information to differentiate similar actions occurring in different spatial contexts. More recently, several datasets
have incorporated road topology information into actions as finer-grained traffic activities.
ROAD~\cite{singh2022road} proposes triplets by imbuing actor labels with traditional actions labels and location-specific annotations for action detection in the spatial domain, while OATS~\cite{agarwal2023ordered} and TACO~\cite{kung2024action} introduce a topology-aware descriptive language that decomposes scenarios into AA and casts the problem as multi-label action recognition.
While these location-aware traffic datasets~\cite{singh2022road, agarwal2023ordered, kung2024action} have refined descriptive and structural atomic activity labels, they primarily rely on egocentric dashcam footage, as shown in Table~\ref{table:dataset}.
This restricts video data to the ego vehicle’s perspective, limiting the ability to analyze broader traffic interactions for important applications such as traffic scenario reconstruction and generation~\cite{dosovitskiy2017carla}. In contrast, ATARS is collected from a top-down perspective, providing full-scene visibility of real intersections. Moreover, we offer multi-label AA annotations at both the video and frame levels, facilitating a more comprehensive traffic activity analysis.

\subsection{UAV-based Traffic Datasets}
Existing UAV-based datasets, such as UAVDT~\cite{du2018unmanned} and VisDrone~\cite{zhu2021detection}, focus on object detection and tracking and thus often lack long-form annotations that describe the progression of traffic participants' actions over time.
On the other hand, UAV-based datasets that emphasize traffic scenarios, such as INTERACTION~\cite{zhan2019interaction}, InD~\cite{bock2020ind}, SIND~\cite{xu2022drone}, CAROM~\cite{lu2023carom}, and CitySim~\cite{zheng2024citysim}, focus on low-level trajectory prediction and often only provide scenario-level labels for accidents, near-misses, and traffic signals in addition to detailed trajectory information.

In this work, we propose ATARS, the first UAV-based dataset designed for traffic activity understanding. ATARS distinguishes itself as the only UAV-based real-world dataset that explicitly annotates topology-aware interactive behaviors, making it crucial for benchmarking models in traffic atomic action recognition and multi-label temporal atomic activity segmentation for aerial data, as well as for atomic activity analysis.

%% file: chapters/4_methodology.tex
\section{The ATARS Dataset}
\label{sec:method}
\begin{figure*}[t!]
    \centering
    \includegraphics[width=1\linewidth]{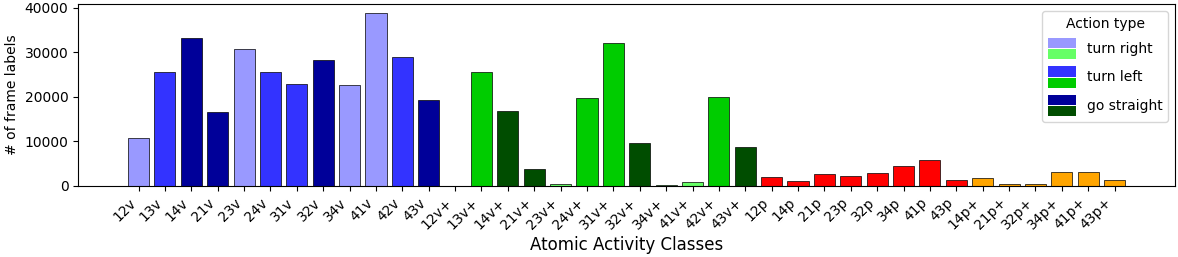}
    \caption{The distribution of atomic activity classes in the ATARS dataset. Four distinct color groups represent the traffic participants \textbf{v}, \textbf{v+}, \textbf{p}, and \textbf{p+}, while different shades for \textbf{v} and \textbf{v+} indicate motion directions: the lightest color denotes right turns, the medium shade denotes going straight, and the darkest shade denotes left turns. Due to the dataset’s long-tail distribution, current models struggle to capture pedestrian patterns. However, this real-world data offers an ideal benchmark for evaluating and improving model robustness in aerial intersection scenarios.}
    \label{fig:distribution}
    \vspace{-5mm}
\end{figure*}
\subsection{Dataset Construction}
We sample untrimmed Full-HD videos from CAROM~\cite{lu2023carom}, a UAV-based dataset containing 69 publicly available videos of nine intersections, to construct our ATARS dataset. For simplicity, we select only the untrimmed videos featuring four-way intersections, excluding those from other intersection types (such as roundabouts) and those with severe camera instability. As a result, ATARS consists of 39 videos recorded at four different four-way intersections. Each video is captured at 30 frames per second, with durations ranging from approximately 45 seconds to 9 minutes. The ATARS dataset is separated into three splits, each containing its own unique roadways. There are 27 videos for the training set, 6 for the validation set, and 6 for the testing set. 

\subsection{Atomic Activity Annotation}
We follow the Atomic Activity (AA) definitions established in the OATS~\cite{agarwal2023ordered} and TACO~\cite{kung2024action} datasets to annotate AA in top-down view interactive traffic scenarios. 
Specifically, each AA contains three annotations: the movement from an entry roadway (\textbf{z1}-\textbf{z4}) or corner (\textbf{c1}-\textbf{c4}) to an exit roadway or corner, and the traffic participant types (\textbf{v}, \textbf{v+}, \textbf{p}, and \textbf{p+}), as shown in Figure~\ref{fig:teaser} (a). 

Two major difference of AA definition between ours and egocentric datasets~\cite{agarwal2023ordered,kung2024action} are made in this work.

\noindent \textbf{Road Topology}: we define road topology regions with intersection-centric instead of egocentric where \textbf{z1} denotes the roadway where the ego-vehicle start moving. In contrast, we define \textbf{z1} as the roadway at the bottom of the image.

\noindent \textbf{Per Frame AA Annotation}:
Unlike prior work that only annotate AA for short clips, we annotate AA for each frame. 
Specifically, the start frame of a vehicle's AA is defined as when the front of the vehicle enters the intersection bypassing the outer line of a crosswalk, or when the vehicle resumes motion after coming to a stop. 
An end frame is then marked when the vehicle completely exits the intersection. 
For pedestrian activities, the start frame is recorded when they exit a corner and enter the intersection, while the end frame is marked when they leave the crosswalk by reaching the goal corner.
The example in Figure~\ref{fig:teaser} (a) shows a group of vehicles moving from \textbf{z1} to \textbf{z3}, which are annotated together as \textbf{13v+}, with `+' indicating the presence of multiple vehicles. While once all but one vehicle exited the intersection, we annotate \textbf{13v}.

Note that in cases where a turning vehicle moves forward into the intersection, stops, and then initiates its turning maneuver into its destination roadway, the previous forward motion is not annotated. 
Instead, the start frame is adjusted to when the turning action begins, as only moving forward is not an indicator of a turning pattern. These stopping periods are also not annotated as positive labels since no motion occurs. 
To this end, we annotated 253,000 atomic activity frames in total. The atomic activity class distribution is illustrated in Figure~\ref{fig:distribution}.

%% file: chapters/5_experiments.tex
\section{Experiments}
\label{sec:exp}

\subsection{Implementation Details}
All models are trained on NVIDIA GeForce RTX 3090 GPU. The model was optimized using the AdamW optimizer and trained for a maximum of 100 epochs to ensure convergence. The learning rate was set to $10^{-4}$. 

\noindent \textbf{Multi-label AA Recognition.} The batch size was set to eight. The input sequence length for all models is 32 frames, which are evenly sampled from a 90-frame video sampled from the untrimmed videos. 
Short videos are sampled continuously from untrimmed videos, without gaps or overlaps. 
To ensure clip quality. we exclude intervals where camera motion is highly unstable. Additionally, we also ensure each clip captures an atomic activity for at least 15 frames. 
In total, 1,009 short videos were sampled. 
For benchmarking, all 40 possible classes were used, including 12 singular and 12 grouped vehicle AA classes, and eight singular and eight grouped pedestrian AA classes.

\noindent \textbf{Multi-label Temporal AA Segmentation.} The batch size is set to two. 
The input for multi-label temporal AA Segmentation is two-stream I3D features~\cite{carreira2017quo} from untrimmed videos, with all 39 untrimmed videos used.
Due to the scarcity of some AA classes, singular and grouped pedestrian classes are combined, resulting in a total of 32 classes being used for the task. To mitigate the long-tail distribution issue, as shown in Figure~\ref{fig:distribution}, we apply data augmentation techniques to the untrimmed videos. 
Specifically, we apply video rotation and flipping and atomic activity labels adjust automatically with rotation and flipping, which is infeasible for egocentric datasets~\cite{agarwal2023ordered,kung2024action} due to their perspective.
Despite these efforts, pedestrian labels remain notably rare and small, posing an ongoing challenge. As shown in Tables~\ref{table:task1} and~\ref{table:task2}, all benchmarking models failed to detect pedestrians, leading to low pedestrian Mean Average Precision (mAP).

Lastly, we use a linear layer that outputs $N_{\mathtt{cl}}$ class channels for multi-label prediction in both tasks. All Models are trained with binary cross-entropy loss (positive weight = 2) and evaluated using video- or frame-level mAP.

\subsection{Baselines}
\label{subsec: baselines}
\noindent \textbf{AA Recognition.} We follow prior work~\cite{kung2024action} for all AA recognition models implementation. 
Specifically, we evaluate various models, including video-level ConvNet: Slowfast~\cite{feichtenhofer2019slowfast} and X3D~\cite{feichtenhofer2020x3d}, transformer-based models: MViT~\cite{fan2021multiscale}, Vivit~\cite{arnab2021vivit}, and VideoMAE~\cite{tong2022videomae}, and object-aware models that use detect-classify paradigm, including ORN~\cite{baradel2018object} and ARG~\cite{wu2019learning}, and the state-of-the-art-model: Action-Slot~\cite{kung2024action}. 
During training, we freeze the backbones and only train the classifiers. 
For object-aware models, we use UAV object detector~\cite{dota-oiell_dataset} to obtain bounding boxes as input. Additionally, we utilize the UAV image segmentation model, SeMask~\cite{jain2023semask}, to obtain background masks, which are used to supervise Action-slot~\cite{kung2024action}. We also implement an LSTM model~\cite{hochreiter1997long} that takes ground-truth object tracking as input.

\noindent \textbf{Temporal AA Segmentation.} We re-implement two categories of existing temporal action segmentation models, including single-label methods designed for capturing multi-scale temporal relationships, including MSTCN~\cite{farha2019ms}, MSTCN++~\cite{li2020ms}, and ASformer~\cite{yi2021asformer}, and multi-label methods that explicitly incorporate modeling activity dependencies, such as PointTAD~\cite{tan2022pointtad}.


\begin{table}[t]
\centering
\small
\caption{Quantitative results on the ATARS dataset. \textbf{v} and \textbf{v+} denote mAP of singular and multiple vehicles activities, while \textbf{p}, and \textbf{p+} denote mAP of singular and multiple pedestrian activities. Best results are in \textbf{bold}, while second-best are \underline{underlined}.
}
\resizebox{0.95\columnwidth}{!}{
\begin{tabular}{@{}l@{\;}c | c @{\;} c@{\;} c@{\;}  c@{\;} |c}
\toprule
\multicolumn{1}{l}{Method} & \multicolumn{1}{c|}{Para. (M)} & \multicolumn{1}{c}{\textbf{v}} & \multicolumn{1}{c}{\textbf{v+}} & \multicolumn{1}{c}{\textbf{p}} & \multicolumn{1}{c}{\textbf{p+}} & \multicolumn{1}{|c}{mAP} \\
\midrule
LSTM w/ GT~\cite{hochreiter1997long} & 0.7  &  0.207 &  0.108 & - & - & -\\
\midrule
SlowFast~\cite{feichtenhofer2019slowfast}  & 33.7 & \textbf{0.767} & \textbf{0.725} & \underline{0.037} &  0.006 & \textbf{0.456}\\
X3D~\cite{feichtenhofer2020x3d} & 3.0 & 0.474 & 0.464 & 0.029 &  0.004 &  0.288 \\
\midrule
MViT~\cite{fan2021multiscale}  & 36.6 & \underline{0.654} & \underline{0.651} & \textbf{0.040} & 0.006 & \underline{0.401} \\
ViviT~\cite{arnab2021vivit}  & 36.6 & 0.212 &  0.108 & 0.036 & 0.006 & 0.110 \\
VideoMAE~\cite{tong2022videomae}  & 57.9 & 0.296 &  0.228 & 0.032 & 0.006 & 0.164 \\
\midrule
ORN~\cite{baradel2018object} & 4.8 & 0.353 & 0.276 & \underline{0.037} & \underline{0.007} & 0.198 \\
ARG~\cite{wu2019learning} & 12.2 & 0.329 & 0.247 & 0.036 & 0.005 & 0.180 \\
\midrule
Action-slot~\cite{kung2024action} & 2.3 & 0.582 & 0.630 & 0.031 & \textbf{0.009} & 0.372 \\
\bottomrule
\end{tabular}
\label{table:task1}
\vspace{-5mm}
}
\end{table}

\subsection{Atomic Activity Recognition Benchmark}
In Table~\ref{table:task1}, we evaluate existing action recognition models. 
We first assess LSTM, which takes ground-truth object tracking as input. Note that we exclude the performance of pedestrians' activities for LSTM due to the lack of pedestrians tracking labels. The inferior performance of LSTM suggests that visual representation is indispensable for the task.
Among the first two groups, models incorporating multi-scale mechanisms, such as SlowFast and MViT, perform significantly better than other video-level models. 
This suggests that our dataset, characterized with small object sizes and varying time spans, requires more robust and diverse levels of spatial-temporal modeling compared to egocentric AA datasets~\cite{agarwal2023ordered,kung2024action}.
While object-aware models, i.e., ORN and ARG, present lower performance because they rely on object detection models, which exacerbates the error accumulation, especially for small objects, such as pedestrians (\textbf{p} and \textbf{p+}) in our dataset. 
Although Action-slot achieves state-of-the-art performance by jointly modeling agent and contextual features on egocentric AA datasets~\cite{agarwal2023ordered,kung2024action}, it presents inferior performance on our UAV dataset. 
We hypothesize that this may be due to the insufficient resolutions of X3D features used as input. This suggests further investigation into integrating multi-scale features into Action-slot.


\subsection{Temporal Atomic Activity Segmentation Benchmark}
In Table~\ref{table:task2}, we evaluate single-label methods such as MS-TCN~\cite{farha2019ms}, MS-TCN++~\cite{li2020ms}, ASFormer~\cite{yi2021asformer}, and multi-label methods, including PointTAD~\cite{tan2022pointtad} for temporal AA segmentation. Interestingly, the overall results remain consistent regardless of whether a method is initially designed for single-label or multi-label temporal action segmentation.
This indicates that existing models struggle to effectively generalize to the challenges in ATARS, particularly the small object sizes.
While the models do yield some correct AA segments for vehicles, they generally fail at detecting pedestrians, which are much smaller. Among the evaluated models, ASformer performs the best. 
Despite PointTAD's massive number of trainable parameters, it still underperformed compared to all evaluated single-label models. We hypothesized that this is due to the data-hungry nature of high parameters and the unpredictability of atomic activity occurrences. 
While patterns of straight moving and cross-turning vehicles have a predictable and repeating pattern due to their movement being dictated by the traffic lights, as seen in Figure~\ref{fig:task2_vis}, the patterns of free-turning vehicles are not, as seen in Figure~\ref{fig:teaser}. 
We also observe that although PointTAD achieves the best in the singular vehicle activities, i.e., \textbf{v}, it performed poorly in grouped vehicles' activities relative to other methods. We hypothesize that the other models generalized better in grouped vehicle scenarios because PointTAD's high-parameter design suffers more from the dataset's long-tail distribution. 
This is also evident from the poorer pedestrian performance relative to other models, as seen in Table~\ref{table:task2}.

\begin{table}[h]
\centering
\small
\caption{Quantitative results on the ATARS dataset. \textbf{v} and \textbf{v+} denote mAP of singular and multiple vehicles activities, while \textbf{p} denote mAP of all pedestrian activities. Best results are in \textbf{bold}, while second-best are \underline{underlined}.
}
\resizebox{0.9\columnwidth}{!}{
\begin{tabular}{@{}l@{\;}c | c@{\;} c@{\;}  c@{\;} | c }
\toprule
\multicolumn{1}{l}{Method} & \multicolumn{1}{c|}{Para. (M)} & \multicolumn{1}{c}{\textbf{v}} & \multicolumn{1}{c}{\textbf{v+}}  & \multicolumn{1}{c}{\textbf{p}} & \multicolumn{1}{|c}{mAP} \\
\midrule
MSTCN~\cite{farha2019ms} & 0.8 & 0.240 & \underline{0.548} & 0.038 & \underline{0.305} \\
MSTCN++~\cite{li2020ms} & 1.0 & 0.227 & \textbf{0.552} & \underline{0.042} & 0.303 \\
ASformer~\cite{yi2021asformer} & 1.1 & \underline{0.306} & 0.531  & \textbf{0.051} & \textbf{0.326} \\
\midrule
PointTAD~\cite{tan2022pointtad} & 210.1 & \textbf{0.346} & 0.329 & 0.002 & 0.253 \\
\bottomrule
\end{tabular}
\label{table:task2}
\vspace{-5mm}
}
\end{table}

The findings from both tasks highlight the challenge of small object size. The results for both AA recognition and multi-label temporal AA segmentation suggest that a wide range of existing methods are equally ineffective at segmenting traffic actions from an aerial view, underscoring the importance of small object detection in this task. These insights underscore the value of ATARS in benchmarking and improving AA recognition and segmentation models in traffic environments.

%% file: chapters/6_discussion.tex
\section{Discussion}

\subsection{Ground-truth vs. Predicted Bounding Boxes}
We study the accumulated error caused by the object detector for object-aware models. Table~\ref{table:orn_arg} reveals that accurate detection is crucial for this type of model. However, even though the aerial view offers relatively accurate calibration for bounding box trajectory, the performance of both models using ground-truth tracking results is still inferior compared to others in Table~\ref{table:task1}, which we assume is due to two reasons. First, object-aware models suffer from associating contextual information, i.e., road topology, with object features, as suggested in prior work~\cite{kung2024action}. Second, the extremely small objects can hinder visual object-feature extraction~\cite{he2017mask}. This result highlights our ATARS dataset's unique challenge that is not presented in existing human activity recognition datasets~\cite{soomro2012ucf101,goyal2017something,kay2017kinetics,gu2018ava,yeung2018every,sigurdsson2016hollywood}.


\begin{table}[h!]
\centering
\small
\caption{Quantitative results of utilizing ground truth (GT) vs. predicted bounding boxes on Temporal Atomic Activity Segmentation.
}
\resizebox{0.8\columnwidth}{!}{
\begin{tabular}{@{} l@{\;}| c@{\;} c @{\;} c@{\;} c@{\;} | c@{\;}}
\toprule
\multicolumn{1}{c}{Method} & \multicolumn{1}{|c}{v} & \multicolumn{1}{c}{v+} & \multicolumn{1}{c}{p} & \multicolumn{1}{c|}{p+} & \multicolumn{1}{c}{mAP} \\
\midrule
ORN~\cite{baradel2018object} & 0.353 & 0.276 & 0.037 & \textbf{0.007} & 0.198 \\
ORN (GT)~\cite{baradel2018object} & \textbf{0.603} &  \textbf{0.640} & \textbf{0.049} & 0.004 & \textbf{0.383} \\

\midrule
ARG~\cite{wu2019learning} &  0.329 & 0.247 & 0.036 & 0.005 & 0.180 \\
ARG (GT)~\cite{wu2019learning} & 0.500 &  0.474 & 0.046 &  \textbf{0.007} & 0.302 \\
\bottomrule
\end{tabular}
\label{table:orn_arg}
\vspace{-5mm}
}
\end{table}

\subsection{Repeating Atomic Activities}
Our ATARS dataset presents unique repetition patterns to traffic lights, as shown in Figure~\ref{fig:task2_vis}. The straight moving vehicle patterns repeat in cycles due to traffic lights.
However, none of the existing methods have the mechanism to exploit this pattern. 
While PointTAD has learnable query points to learn where an action generally occurs through training, these query points are fixed during inference. 
Fixed query points are unable to fully leverage the cyclic behaviors of intersections because the presence, phase shifts, and duration of their turn takings (frequency) for traffic light cycles with different intersections in different videos are not the same. 
Hence, fixed query points learned from phase shifts and frequencies of training data cannot generalize well to describe phase shift and frequency of a different intersection in a different video. 
A promising direction is to employ test-time adaptable query points, which can recognize the effect of traffic lights and adjust query points accordingly. 
Another promising direction is to incorporate additional infrastructure data to model the repeating patterns, such as traffic lights and the duration of their turn takings. 
We hope this insight will promote additional future joint efforts to study multi-modal representation learning that incorporates infrastructure sensor data for traffic analysis. 

\subsection{Sparse and Short Action Segments}
\begin{figure}[t!]
    \centering
    \includegraphics[width=1\linewidth]{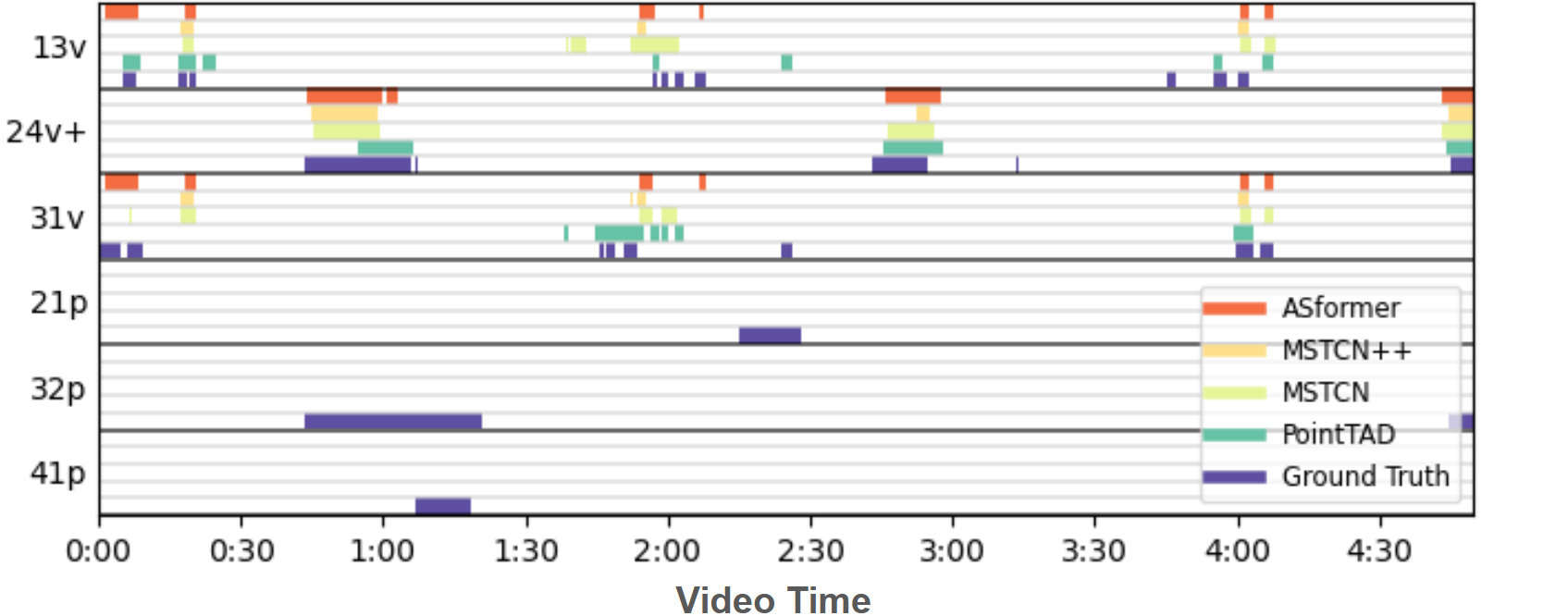}
    \caption{Qualitative Visualization of various methods in multi-label temporal atomic activity segmentation~\cite{farha2019ms,li2020ms,yi2021asformer}. The dark indigo segment represents the ground truth, and the purple segment denotes correct predictions that align with the ground truth.}
    \label{fig:task2_vis}
    \vspace{-5mm}
\end{figure}

Our findings highlight the difficulty that the evaluated methods face in detecting and segmenting short segments. Fast-moving or quick-turning cars leave short and spontaneous segments, as can be seen in 
Figure~\ref{fig:teaser} and Figure~\ref{fig:task2_vis}. 
Existing work~\cite{yi2021asformer,tan2022pointtad} in temporal action segmentation is fixated on competing on the long segment benchmark dataset~\cite{yeung2018every,sigurdsson2016hollywood}. 
They proposed various mechanisms~\cite {yi2021asformer} for resolving over-segment problems while neglecting the need to handle short segments. 
A promising direction is the employment of multi-temporal mechanisms and class adaptive temporal attention. 
Another issue related to the sparse segments worth noting is that, our multi-label temporal AA segmentation task further exacerbates the long-tail issue as it expands the imbalance between frequent and infrequent actions. Unlike standard single-label temporal segmentation, where rare actions are already challenging to detect, the multi-label nature of our task increases the likelihood of under-representing small and sparse segments. 
To address this, future research should consider leveraging adaptive temporal modeling techniques that explicitly balance long and short segments.

\subsection{Small Traffic Participant Detection}
Our findings highlight the difficulty state-of-the-art methods face in detecting and segmenting small objects, particularly pedestrians, in aerial traffic scenes. Despite the video being taken at high resolution, pedestrians are only several pixels wide in a Full-HD video. As shown in Table~\ref{table:task1}, Table~\ref{table:task2}, and Figure~\ref{fig:task2_vis}, existing models are unable to capture these small pedestrians while overlooking the large intersection, and this problem is worsened in models designed with down-scaling or low-resolution mechanisms, such as slots. A promising direction is the development of methods that focus on multi-scale up-sampling feature aggregation, super-resolution pre-processing, and high-resolution spatiotemporal attention mechanisms to enhance small-object detection. 


\subsection{Atomic Activity Interaction}
We evaluate the models that explicitly learn the interaction between atomic activities, i.e., object-aware models ORN and ARG that utilize GCN to model the relationships of objects in AA recognition in Table~\ref{table:task1} and models focus on multi-label temporal action segmentation, such as PointTAD in Table~\ref{table:task2}. 
We further conduct experiments to enhance the Action-slot's ability to model atomic activity interaction by adding an MLP to the slots that capture corresponding classes of atomic activity. 
However, the performance drops significantly as shown in Table~\ref{table:as_mlp}.
We found out that these effective and popular mechanisms used in existing human action datasets~\cite{yeung2018every,sigurdsson2016hollywood} surprisingly perform worse on our dataset and tasks. We assume this may be due to traffic participants' behavior being highly constrained by traffic regularization such as traffic lights instead of the behavior of each other.

\begin{table}[h!]
\centering
\small
\caption{Quantitative results of Action-Slot with and w/o MLP on Multi-label Atomic Activity Recognition.
}
\resizebox{0.9\columnwidth}{!}{
\begin{tabular}{@{} l@{\;}| c@{\;} c @{\;} c@{\;} c@{\;} | c@{\;}}
\toprule
\multicolumn{1}{c}{Method} & \multicolumn{1}{|c}{v} & \multicolumn{1}{c}{v+} & \multicolumn{1}{c}{p} & \multicolumn{1}{c|}{p+} & \multicolumn{1}{c}{mAP} \\
\midrule

Action-slot~\cite{kung2024action} & \textbf{0.582} & \textbf{0.630} & 0.031 & \textbf{0.009} & \textbf{0.372} \\
Action-slot~\cite{kung2024action} w/ MLP & 0.539 &  0.343 &  \textbf{0.034} & 0.004 & 0.272 \\
\bottomrule
\end{tabular}
\label{table:as_mlp}
\vspace{-5mm}
}
\end{table}

%% file: chapters/7_conclusion.tex
\section{Conclusion}
\label{sec:con}
Our ATARS dataset creates new opportunities for studying traffic activity from an aerial view, which is crucial for many applications that egocentric datasets cannot support. We offer both video- and frame-level atomic activity labels and introduce a new task, multi-label temporal atomic activity recognition to enable finer-grained traffic flow analysis. 
Through extensive evaluations of SOTA models, we provide a comprehensive analysis of their limitations on our dataset, including challenges with repeating atomic activities~\cite{tan2022pointtad}, sparse and short segments~\cite{yi2021asformer}, and large size variations of traffic participants~\cite{kung2024action}. 
These factors caused existing methods to fail to perform well on aerial data.
We hope our dataset and insightful findings will facilitate the development of more advanced traffic activity modeling methodologies and strategies specifically designed for aerial atomic activities.

\noindent \textbf{Future Work.}
Expanding the ATARS dataset is crucial for advancing traffic action analysis. Annotating traffic lights would enrich contextual modeling. Including diverse intersection types, such as roundabouts and multi-lane merges, would enhance model applicability to varied urban settings.

%% file: main.bbl
\begin{thebibliography}{10}
\providecommand{\url}[1]{#1}
\csname url@rmstyle\endcsname
\providecommand{\newblock}{\relax}
\providecommand{\bibinfo}[2]{#2}
\providecommand\BIBentrySTDinterwordspacing{\spaceskip=0pt\relax}
\providecommand\BIBentryALTinterwordstretchfactor{4}
\providecommand\BIBentryALTinterwordspacing{\spaceskip=\fontdimen2\font plus
\BIBentryALTinterwordstretchfactor\fontdimen3\font minus \fontdimen4\font\relax}
\providecommand\BIBforeignlanguage[2]{{%
\expandafter\ifx\csname l@#1\endcsname\relax
\typeout{** WARNING: IEEEtran.bst: No hyphenation pattern has been}%
\typeout{** loaded for the language `#1'. Using the pattern for}%
\typeout{** the default language instead.}%
\else
\language=\csname l@#1\endcsname
\fi
#2}}

\bibitem{singh2022road}
G.~Singh, S.~Akrigg, M.~Di~Maio, V.~Fontana, R.~J. Alitappeh, S.~Khan, S.~Saha, K.~Jeddisaravi, F.~Yousefi, J.~Culley, \emph{et~al.}, ``Road: The road event awareness dataset for autonomous driving,'' \emph{IEEE Trans. Pattern Anal.Mach. Intell.}, 2022.

\bibitem{malla2023drama}
S.~Malla, C.~Choi, I.~Dwivedi, J.~H. Choi, and J.~Li, ``Drama: Joint risk localization and captioning in driving,'' in \emph{Proceedings of the IEEE/CVF winter conference on applications of computer vision}, 2023.

\bibitem{ramanishka2018toward}
V.~Ramanishka, Y.-T. Chen, T.~Misu, and K.~Saenko, ``Toward driving scene understanding: A dataset for learning driver behavior and causal reasoning,'' in \emph{Proc. Conf. Comput. Vis. Pattern Recognit.}, 2018.

\bibitem{agarwal2023ordered}
N.~Agarwal and Y.-T. Chen, ``Ordered atomic activity for fine-grained interactive traffic scenario understanding,'' in \emph{Proc. Int. Conf. Comput. Vis.}, 2023.

\bibitem{kung2024action}
C.-H. Kung, S.-W. Lu, Y.-H. Tsai, and Y.-T. Chen, ``Action-slot: Visual action-centric representations for multi-label atomic activity recognition in traffic scenes,'' in \emph{Proc. Conf. Comput. Vis. Pattern Recognit.}, 2024.

\bibitem{du2018unmanned}
D.~Du, Y.~Qi, H.~Yu, Y.~Yang, K.~Duan, G.~Li, W.~Zhang, Q.~Huang, and Q.~Tian, ``The unmanned aerial vehicle benchmark: Object detection and tracking,'' in \emph{Proc. European Conf. Computer Vis.}, 2018.

\bibitem{zhu2021detection}
P.~Zhu, L.~Wen, D.~Du, X.~Bian, H.~Fan, Q.~Hu, and H.~Ling, ``Detection and tracking meet drones challenge,'' \emph{IEEE Trans. Pattern Anal.Mach. Intell.}, 2021.

\bibitem{zheng2024citysim}
O.~Zheng, M.~Abdel-Aty, L.~Yue, A.~Abdelraouf, Z.~Wang, and N.~Mahmoud, ``Citysim: A drone-based vehicle trajectory dataset for safety-oriented research and digital twins,'' \emph{Transportation research record}, 2024.

\bibitem{zhan2019interaction}
W.~Zhan, L.~Sun, D.~Wang, H.~Shi, A.~Clausse, M.~Naumann, J.~Kummerle, H.~Konigshof, C.~Stiller, A.~de~La~Fortelle, \emph{et~al.}, ``Interaction dataset: An international, adversarial and cooperative motion dataset in interactive driving scenarios with semantic maps,'' \emph{arXiv preprint arXiv:1910.03088}, 2019.

\bibitem{bock2020ind}
J.~Bock, R.~Krajewski, T.~Moers, S.~Runde, L.~Vater, and L.~Eckstein, ``The ind dataset: A drone dataset of naturalistic road user trajectories at german intersections,'' in \emph{2020 IEEE Intelligent Vehicles Symposium (IV)}, 2020.

\bibitem{xu2022drone}
Y.~Xu, W.~Shao, J.~Li, K.~Yang, W.~Wang, H.~Huang, C.~Lv, and H.~Wang, ``Sind: A drone dataset at signalized intersection in china,'' in \emph{2022 IEEE 25th International Conference on Intelligent Transportation Systems (ITSC)}, 2022.

\bibitem{lu2023carom}
D.~Lu, E.~Eaton, M.~Weg, W.~Wang, S.~Como, J.~Wishart, H.~Yu, and Y.~Yang, ``Carom air-vehicle localization and traffic scene reconstruction from aerial videos,'' in \emph{2023 IEEE International Conference on Robotics and Automation (ICRA)}, 2023.

\bibitem{lin2014visual}
D.~Lin, S.~Fidler, C.~Kong, and R.~Urtasun, ``Visual semantic search: Retrieving videos via complex textual queries,'' in \emph{Proceedings of the IEEE conference on computer vision and pattern recognition}, 2014.

\bibitem{segal2021universal}
S.~Segal, E.~Kee, W.~Luo, A.~Sadat, E.~Yumer, and R.~Urtasun, ``Universal embeddings for spatio-temporal tagging of self-driving logs,'' in \emph{Conference on Robot Learning}, 2021.

\bibitem{naphade20237th}
M.~Naphade, S.~Wang, D.~C. Anastasiu, Z.~Tang, M.-C. Chang, Y.~Yao, L.~Zheng, M.~S. Rahman, M.~S. Arya, A.~Sharma, \emph{et~al.}, ``The 7th ai city challenge,'' in \emph{Proceedings of the IEEE/CVF Conference on Computer Vision and Pattern Recognition}, 2023.

\bibitem{taha2020boosting}
A.~Taha, Y.-T. Chen, T.~Misu, A.~Shrivastava, and L.~Davis, ``Boosting standard classification architectures through a ranking regularizer,'' in \emph{Proceedings of the IEEE/CVF Winter Conference on Applications of Computer Vision}, 2020.

\bibitem{cao2022advdo}
Y.~Cao, C.~Xiao, A.~Anandkumar, D.~Xu, and M.~Pavone, ``Advdo: Realistic adversarial attacks for trajectory prediction,'' in \emph{Proc. European Conf. Computer Vis.}, 2022.

\bibitem{Rempe_2022_CVPR}
D.~Rempe, J.~Philion, L.~J. Guibas, S.~Fidler, and O.~Litany, ``Generating useful accident-prone driving scenarios via a learned traffic prior,'' in \emph{Proc. Conf. Comput. Vis. Pattern Recognit.}, 2022.

\bibitem{Rempe_2023_CVPR}
D.~Rempe, Z.~Luo, X.~Bin~Peng, Y.~Yuan, K.~Kitani, K.~Kreis, S.~Fidler, and O.~Litany, ``Trace and pace: Controllable pedestrian animation via guided trajectory diffusion,'' in \emph{Proceedings of the IEEE/CVF Conference on Computer Vision and Pattern Recognition (CVPR)}, 2023.

\bibitem{NEURIPS2022_a48ad12d}
C.~Xu, W.~Ding, W.~Lyu, Z.~LIU, S.~Wang, Y.~He, H.~Hu, D.~ZHAO, and B.~Li, ``Safebench: A benchmarking platform for safety evaluation of autonomous vehicles,'' in \emph{Proc. Neural Inf. Process. Syst.}, 2022.

\bibitem{make_stop}
C.~Li, S.~H. Chan, and Y.-T. Chen, ``Who make drivers stop? towards driver-centric risk assessment: Risk object identification via causal inference,'' in \emph{2020 IEEE/RSJ International Conference on Intelligent Robots and Systems (IROS)}, 2020.

\bibitem{droid}
------, ``Droid: Driver-centric risk object identification,'' \emph{IEEE Transactions on Pattern Analysis and Machine Intelligence}, 2023.

\bibitem{riskbench}
C.-H. Kung, C.-C. Yang, P.-Y. Pao, S.-W. Lu, P.-L. Chen, H.-C. Lu, and Y.-T. Chen, ``Riskbench: A scenario-based benchmark for risk identification,'' in \emph{2024 IEEE International Conference on Robotics and Automation (ICRA)}, 2024.

\bibitem{pao2024potential}
P.-Y. Pao, S.-W. Lu, Z.-Y. Lu, and Y.-T. Chen, ``Potential field as scene affordance for behavior change-based visual risk object identification,'' \emph{arXiv preprint arXiv:2409.15846}, 2024.

\bibitem{chen2016atomic}
C.-Y. Chen, W.~Choi, and M.~Chandraker, ``Atomic scenes for scalable traffic scene recognition in monocular videos,'' in \emph{2016 IEEE Winter Conference on Applications of Computer Vision (WACV)}, 2016.

\bibitem{li2019dbus}
M.~G. Li, B.~Jiang, Z.~Che, X.~Shi, M.~Liu, Y.~Meng, J.~Ye, and Y.~Liu, ``Dbus: Human driving behavior understanding system,'' in \emph{2019 IEEE/CVF International Conference on Computer Vision Workshop (ICCVW)}.\hskip 1em plus 0.5em minus 0.4em\relax IEEE, 2019, pp. 2436--2444.

\bibitem{malla2020titan}
S.~Malla, B.~Dariush, and C.~Choi, ``Titan: Future forecast using action priors,'' in \emph{Proceedings of the IEEE/CVF Conference on Computer Vision and Pattern Recognition}, 2020.

\bibitem{hu2024optimizing}
Y.~Hu, D.~Rey, R.~Mohajerpoor, and M.~Saberi, ``Optimizing traffic signal control for continuous-flow intersections: Benchmarking against a state-of-practice model,'' \emph{IET Intelligent Transport Systems}, 2024.

\bibitem{dosovitskiy2017carla}
A.~Dosovitskiy, G.~Ros, F.~Codevilla, A.~Lopez, and V.~Koltun, ``Carla: An open urban driving simulator,'' in \emph{Conference on robot learning}, 2017.

\bibitem{baradel2018object}
F.~Baradel, N.~Neverova, C.~Wolf, J.~Mille, and G.~Mori, ``Object level visual reasoning in videos,'' in \emph{Proc. European Conf. Computer Vis.}, 2018.

\bibitem{wu2019learning}
J.~Wu, L.~Wang, L.~Wang, J.~Guo, and G.~Wu, ``Learning actor relation graphs for group activity recognition,'' in \emph{Proc. Conf. Comput. Vis. Pattern Recognit.}, 2019.

\bibitem{feichtenhofer2019slowfast}
C.~Feichtenhofer, H.~Fan, J.~Malik, and K.~He, ``Slowfast networks for video recognition,'' in \emph{Proc. Int. Conf. Comput. Vis.}, 2019.

\bibitem{feichtenhofer2020x3d}
C.~Feichtenhofer, ``X3d: Expanding architectures for efficient video recognition,'' in \emph{Proc. Conf. Comput. Vis. Pattern Recognit.}, 2020.

\bibitem{fan2021multiscale}
H.~Fan, B.~Xiong, K.~Mangalam, Y.~Li, Z.~Yan, J.~Malik, and C.~Feichtenhofer, ``Multiscale vision transformers,'' in \emph{Proc. Int. Conf. Comput. Vis.}, 2021.

\bibitem{arnab2021vivit}
A.~Arnab, M.~Dehghani, G.~Heigold, C.~Sun, M.~Lu{\v{c}}i{\'c}, and C.~Schmid, ``Vivit: A video vision transformer,'' in \emph{Proc. Int. Conf. Comput. Vis.}, 2021.

\bibitem{tong2022videomae}
Z.~Tong, Y.~Song, J.~Wang, and L.~Wang, ``Video{MAE}: Masked autoencoders are data-efficient learners for self-supervised video pre-training,'' in \emph{Proc. Neural Inf. Process. Syst.}, 2022.

\bibitem{farha2019ms}
Y.~A. Farha and J.~Gall, ``Ms-tcn: Multi-stage temporal convolutional network for action segmentation,'' in \emph{Proc. Conf. Comput. Vis. Pattern Recognit.}, 2019, pp. 3575--3584.

\bibitem{li2020ms}
S.~Li, Y.~A. Farha, Y.~Liu, M.-M. Cheng, and J.~Gall, ``Ms-tcn++: Multi-stage temporal convolutional network for action segmentation,'' \emph{IEEE Trans. Pattern Anal.Mach. Intell.}, 2020.

\bibitem{yi2021asformer}
F.~Yi, H.~Wen, and T.~Jiang, ``Asformer: Transformer for action segmentation,'' \emph{arXiv preprint arXiv:2110.08568}, 2021.

\bibitem{tan2022pointtad}
J.~Tan, X.~Zhao, X.~Shi, B.~Kang, and L.~Wang, ``Pointtad: Multi-label temporal action detection with learnable query points,'' \emph{Proc. Neural Inf. Process. Syst.}, 2022.

\bibitem{geiger2013vision}
A.~Geiger, P.~Lenz, C.~Stiller, and R.~Urtasun, ``Vision meets robotics: The kitti dataset,'' \emph{The international journal of robotics research}, 2013.

\bibitem{yu2020bdd100k}
F.~Yu, H.~Chen, X.~Wang, W.~Xian, Y.~Chen, F.~Liu, V.~Madhavan, and T.~Darrell, ``Bdd100k: A diverse driving dataset for heterogeneous multitask learning,'' in \emph{Proc. Conf. Comput. Vis. Pattern Recognit.}, 2020.

\bibitem{caesar2020nuscenes}
H.~Caesar, V.~Bankiti, A.~H. Lang, S.~Vora, V.~E. Liong, Q.~Xu, A.~Krishnan, Y.~Pan, G.~Baldan, and O.~Beijbom, ``nuscenes: A multimodal dataset for autonomous driving,'' in \emph{Proceedings of the IEEE/CVF conference on computer vision and pattern recognition}, 2020.

\bibitem{chang2019argoverse}
M.-F. Chang, J.~Lambert, P.~Sangkloy, J.~Singh, S.~Bak, A.~Hartnett, D.~Wang, P.~Carr, S.~Lucey, D.~Ramanan, \emph{et~al.}, ``Argoverse: 3d tracking and forecasting with rich maps,'' in \emph{Proceedings of the IEEE/CVF conference on computer vision and pattern recognition}, 2019.

\bibitem{sun2020scalability}
P.~Sun, H.~Kretzschmar, X.~Dotiwalla, A.~Chouard, V.~Patnaik, P.~Tsui, J.~Guo, Y.~Zhou, Y.~Chai, B.~Caine, \emph{et~al.}, ``Scalability in perception for autonomous driving: Waymo open dataset,'' in \emph{Proceedings of the IEEE/CVF conference on computer vision and pattern recognition}, 2020.

\bibitem{carreira2017quo}
J.~Carreira and A.~Zisserman, ``Quo vadis, action recognition? a new model and the kinetics dataset,'' in \emph{Proc. Conf. Comput. Vis. Pattern Recognit.}, 2017.

\bibitem{dota-oiell_dataset}
\BIBentryALTinterwordspacing
ObjectDetectionModels, ``Dota dataset,'' 2024. [Online]. Available: \url{https://universe.roboflow.com/objectdetectionmodels/dota-oiell}
\BIBentrySTDinterwordspacing

\bibitem{jain2023semask}
J.~Jain, A.~Singh, N.~Orlov, Z.~Huang, J.~Li, S.~Walton, and H.~Shi, ``Semask: Semantically masked transformers for semantic segmentation,'' in \emph{Proceedings of the IEEE/CVF international conference on computer vision}, 2023, pp. 752--761.

\bibitem{hochreiter1997long}
S.~Hochreiter and J.~Schmidhuber, ``Long short-term memory,'' \emph{Neural computation}, vol.~9, no.~8, pp. 1735--1780, 1997.

\bibitem{he2017mask}
K.~He, G.~Gkioxari, P.~Doll{\'a}r, and R.~Girshick, ``Mask r-cnn,'' in \emph{Proc. Int. Conf. Comput. Vis.}, 2017.

\bibitem{soomro2012ucf101}
K.~Soomro, A.~R. Zamir, and M.~Shah, ``Ucf101: A dataset of 101 human actions classes from videos in the wild,'' \emph{arXiv preprint arXiv:1212.0402}, 2012.

\bibitem{goyal2017something}
R.~Goyal, S.~Ebrahimi~Kahou, V.~Michalski, J.~Materzynska, S.~Westphal, H.~Kim, V.~Haenel, I.~Fruend, P.~Yianilos, M.~Mueller-Freitag, \emph{et~al.}, ``The" something something" video database for learning and evaluating visual common sense,'' in \emph{Proc. Int. Conf. Comput. Vis.}, 2017.

\bibitem{kay2017kinetics}
W.~Kay, J.~Carreira, K.~Simonyan, B.~Zhang, C.~Hillier, S.~Vijayanarasimhan, F.~Viola, T.~Green, T.~Back, P.~Natsev, \emph{et~al.}, ``The kinetics human action video dataset,'' \emph{arXiv preprint arXiv:1705.06950}, 2017.

\bibitem{gu2018ava}
C.~Gu, C.~Sun, D.~A. Ross, C.~Vondrick, C.~Pantofaru, Y.~Li, S.~Vijayanarasimhan, G.~Toderici, S.~Ricco, R.~Sukthankar, \emph{et~al.}, ``Ava: A video dataset of spatio-temporally localized atomic visual actions,'' in \emph{Proc. Conf. Comput. Vis. Pattern Recognit.}, 2018.

\bibitem{yeung2018every}
S.~Yeung, O.~Russakovsky, N.~Jin, M.~Andriluka, G.~Mori, and L.~Fei-Fei, ``Every moment counts: Dense detailed labeling of actions in complex videos,'' \emph{Int. J. Comput. Vis.}, 2018.

\bibitem{sigurdsson2016hollywood}
G.~A. Sigurdsson, G.~Varol, X.~Wang, A.~Farhadi, I.~Laptev, and A.~Gupta, ``Hollywood in homes: Crowdsourcing data collection for activity understanding,'' in \emph{Proc. European Conf. Computer Vis.}, 2016.

\end{thebibliography}
